\pdfoutput=1
\documentclass[11pt]{article}

\usepackage[final]{acl}

\usepackage{times}
\usepackage{latexsym}

\usepackage[T1]{fontenc}
\usepackage[utf8]{inputenc}

\usepackage{microtype}

\usepackage{inconsolata}

\usepackage{graphicx}

\usepackage{amsmath}
\usepackage{extpfeil}
\usepackage{multirow}
\usepackage{booktabs}
\usepackage{tabularx}
\usepackage{adjustbox}
\usepackage{natbib}
\usepackage{pdfpages}
\usepackage{subfigure}
\usepackage{tikz}

\usepackage{pgfplotstable}
\usepackage{booktabs}

\title{ReCUT: Balancing Reasoning Length and Accuracy in LLMs via Stepwise Trails and Preference Optimization}

\author{First Author \\
  Affiliation / Address line 1 \\
  Affiliation / Address line 2 \\
  Affiliation / Address line 3 \\
  \texttt{email@domain} \\\And
  Second Author \\
  Affiliation / Address line 1 \\
  Affiliation / Address line 2 \\
  Affiliation / Address line 3 \\
  \texttt{email@domain} \\}

\author{Zhensheng Jin$^{1}$\thanks{ \ \ indicates equal contribution.}, Xinze Li$^{1}$\footnotemark[1], Yifan Ji$^{1}$, Chunyi Peng$^{1}$, Zhenghao Liu$^{1}$\thanks{ \ \ indicates corresponding author.}, Qi Shi$^{2}$,\\\textbf{Yukun Yan$^{2}$, Shuo Wang$^{2}$, Furong Peng$^{3}$, Ge Yu$^{1}$} \\ 
$^1$ School of Computer Science and Engineering, Northeastern University, China \\
$^2$ Department of Computer Science and Technology, Institute for AI, Tsinghua University, China \\
Beijing National Research Center for Information Science and Technology, China \\
$^3$Institute of Big Data Science and Industry/School of Computer and \\Information Technology, Shanxi University, China\\}

\begin{document}
\maketitle

\begin{abstract}
% Recent advances in Chain-of-Thought (CoT) prompting have significantly enhanced the reasoning capabilities of large language models LLMs. However, these models often exhibit overthinking behaviors, generating unnecessarily long reasoning trajectories that compromise computational efficiency and can even reduce accuracy. 
Recent advances in Chain-of-Thought (CoT) prompting have substantially improved the reasoning capabilities of Large Language Models (LLMs). However, these methods often suffer from overthinking, leading to unnecessarily lengthy or redundant reasoning traces. Existing approaches attempt to mitigate this issue through curating multiple reasoning chains for training LLMs, but their effectiveness is often constrained by the quality of the generated data and prone to overfitting. To address the challenge, we propose \textbf{Re}asoning \textbf{C}ompression Thro\textbf{U}gh Stepwise \textbf{T}rials (ReCUT), a novel method aimed at balancing the accuracy and length of reasoning trajectory. Specifically, ReCUT employs a stepwise exploration mechanism and a long-short switched sampling strategy, enabling LLMs to incrementally generate diverse reasoning paths. These paths are evaluated and used to construct preference pairs to train two specialized models (Gemini LLMs)---one optimized for reasoning accuracy, the other for shorter reasoning. A final integrated model is obtained by interpolating the parameters of these two models. Experimental results across multiple math reasoning datasets and backbone models demonstrate that ReCUT significantly reduces reasoning lengths by approximately 30-50\%, while maintaining or improving reasoning accuracy compared to various baselines. All codes and data will be released via \url{https://github.com/NEUIR/ReCUT}.
\end{abstract}
\section{Introduction}
\begin{figure}[t!]
    \includegraphics[width=\linewidth]{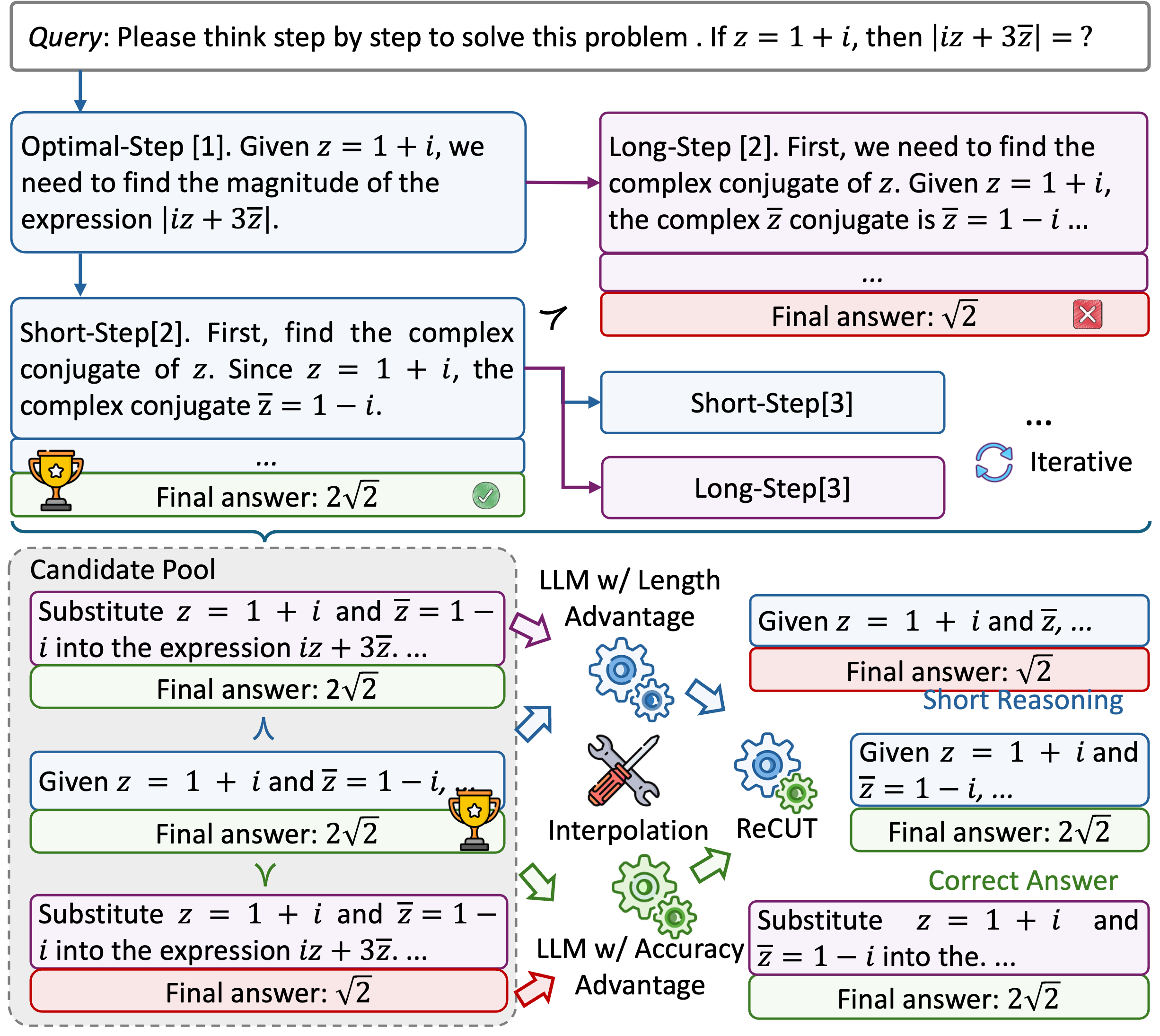}
    \caption{Illustration of Our Reasoning Compression Through Stepwise Trials (ReCUT) Model.}
    \label{fig:into}
\end{figure}
Recent advances in Chain-of-Thought (CoT)~\cite{wei2022chain} have substantially improved the reasoning capabilities of Large Language Models (LLMs), motivating researchers to explore a new scaling paradigm---test-time scaling~\cite{muennighoff2025s1,snell2024scaling}. This paradigm improves the performance of LLMs on lots of challenging reasoning tasks, such as math competitions~\cite{hendrycks2measuring} and PhD-level subject QA~\cite{rein2024gpqa}, by extending the CoT with deeper and iterative thinking during inference~\cite{wu2024empirical,snell2024scaling}. While effective, test-time scaling incurs higher computational costs~\cite{lee2025well}. Furthermore, recent studies suggest that LLMs often exhibit overthinking behaviors, even on relatively simple problems~\cite{chen2024not}, potentially diminishing the benefits of deeper reasoning in certain cases.

To address the inefficiencies caused by overthinking, various strategies have been proposed. Prompt-based methods aim to guide LLMs to generate more concise reasoning trajectories. SFT-based methods prompt LLMs to sample multiple reasoning trajectories and select the concise and correct ones to synthesize SFT data for fine-tuning itself~\cite{team2025kimi}. However, the former may cause the LLMs to omit critical intermediate steps~\cite{jin2024impact,lee2025well}, while the latter may lead to overfitting to the training signals~\cite{luo2023empirical}. Reinforcement learning (RL)-based methods represent another promising research direction~\cite{aggarwal2025l1}, where carefully designed reward functions penalize overly long reasoning trajectories and inaccurate outputs, thereby guiding LLMs to produce concise and accurate reasoning results.

% To address the inefficiencies caused by overthinking, various strategies have been proposed. Prompt-based compression methods aim to guide LLMs to conduct more concise reasoning results using carefully designed instructions~\cite{xu2025chain}. While effective in reducing output length, these methods often cause LLMs to omit critical intermediate steps, resulting in degraded answer quality~\cite{jin2024impact,lee2025well}. SFT-based methods prompt LLMs to sample multiple reasoning trajectories and select the concise and correct ones to synthesize SFT data for fine-tuning itself. However, this approach relies on the model's ability~\cite{qiao2025concise} to synthesize the SFT dataset and lead to overfitting to training signals~\cite{luo2023empirical}. Reinforcement learning (RL)-based methods represent another promising research direction~\cite{aggarwal2025l1}, where carefully designed reward functions penalize overly long reasoning trajectories and inaccurate outputs, thereby guiding LLMs to produce concise and accurate reasoning results. 

Existing RL-based methods typically sample multiple complete reasoning trajectories using a single instruction to compute their rewards and construct preference data for training. However, this sampling strategy inherently limits the diversity of reasoning trajectories~\cite{liu2023statistical}, resulting in convergent distributions in terms of both length and accuracy. Furthermore, such approaches often overlook the fine-grained contributions of individual reasoning steps in reasoning trajectories, allowing even the correct trajectories to include redundant steps~\cite{wang2025stepwise}. These limitations compromise the quality of the constructed preference data, ultimately resulting in suboptimal reasoning performance, which motivates us to explore a better method for constructing high-quality preference data. Recent methods that search for optimal reasoning paths by stepwise decoding have achieved advantages in reducing reasoning errors and redundancies generated by LLMs~\cite{wang2025stepwise}. This success inspires us to construct high-quality and diverse preference data by prompting the model to stepwise explore reasoning paths.

In this paper, we propose ReCUT (Reasoning Compression Through Stepwise Trials), a method that guides LLMs to progressively explore diverse reasoning trajectories and optimizes them via preference-based learning to balance reasoning accuracy and length. As shown in Figure~\ref{fig:into}, ReCUT introduces a stepwise reasoning trajectory exploration mechanism: at each step, the LLM is conditioned on a partially optimal trajectory and generates all subsequent reasoning steps using a long-short sampling strategy to encourage diversity. All trajectories produced via stepwise sampling are incorporated with the given optimal trajectory and then collected into a candidate pool. Then the first step of the generated reasoning outcome is further evaluated to incrementally construct the optimal reasoning path. After trajectory collection, ReCUT leverages this candidate pool to train two specialized LLMs (Gemini LLMs)---one favoring accuracy and the other favoring shorter reasoning results. Finally, ReCUT performs parameter interpolation between these two optimized LLMs to achieve a trade-off between the reasoning accuracy and the length of reasoning results.

Our experiments demonstrate the effectiveness of ReCUT, which significantly reduces the reasoning length, typically by 30-50\%, while maintaining or even surpassing the accuracy of baseline methods. Further analysis reveals that our stepwise sampling strategy enables the synthesis of diverse reasoning trajectories, helping LLMs regulate the number of reasoning steps through preference optimization. By leveraging parameter interpolation, ReCUT not only achieves higher accuracy but also further shortens the reasoning process by combining the strengths of both Gemini LLMs. Moreover, ReCUT effectively mitigates overthinking in LLMs by reducing redundant reasoning steps and identifying unproductive reasoning paths that may lead to incorrect answers.
\section{Related Work}
The test-time scaling law suggests that LLMs, such as Deepseek-R1~\cite{deepseekai2025deepseekr1incentivizingreasoningcapability} and QwQ~\cite{qwq32b}, can enhance their performance on challenging reasoning and mathematical tasks~\cite{wu2024empirical,snell2024scaling} by engaging in deeper reasoning~\cite{deepseekai2025deepseekr1incentivizingreasoningcapability} and producing longer Chains-of-Thought (CoT)~\cite{wei2022chain}. While effective, these approaches often cause LLMs to overthink~\cite{sui2025stop}, resulting in redundant or irrelevant content~\cite{chiang2024over}, which not only increases inference cost but may also lead to incorrect answers~\cite{cuadron2025danger}.

% To better control and reduce the length of CoT generated by LLMs~\cite{lee2025well,aggarwal2025l1,cui2025stepwise},

To better balance reasoning efficiency and accuracy, recent work has explored efficient reasoning with LLMs, with a particular focus on reducing and controlling the length of reasoning chains~\cite{lee2025well,aggarwal2025l1,cui2025stepwise}. Some approaches design specialized prompts to directly shorten the generated responses~\cite{renze2024benefits,lee2025well}. However, such strategies may truncate essential intermediate reasoning steps, ultimately compromising accuracy~\cite{aggarwal2025l1}. To address this problem, \citet{kang2025c3ot,muennighoff2025s1} leverage advanced LLMs (e.g., ChatGPT~\cite{openai2024gpt4technicalreport}) to generate concise yet accurate CoTs as training data for Supervised Fine-Tuning (SFT). Nevertheless, this method is inherently limited by the capabilities of the teacher models and may lead to overfitting to the provided training signals.

Benefiting from advances in reinforcement learning~\cite{schulman2017proximal,shao2024deepseekmath}, recent studies have leveraged it to enable LLMs to adaptively control the length of their reasoning processes. \citet{aggarwal2025l1} introduce a length bias penalty and an accuracy reward during training to encourage LLMs to generate accurate reasoning using fewer tokens. \citet{chen2025towards} propose a cosine reward mechanism that promotes reasoning trajectories of moderate length by penalizing both overly short and overly long generations. However, these approaches compute rewards based on entire reasoning trajectories, without estimating the contribution of individual steps in the whole reasoning chain. In contrast, ReCUT leverages a step-wise sampling strategy to elicit more fine-grained and diverse preference data from LLMs, which enhances the training process for reasoning compression.

\section{Methodology}
\begin{figure*}[t!]
    \includegraphics[width=\textwidth]{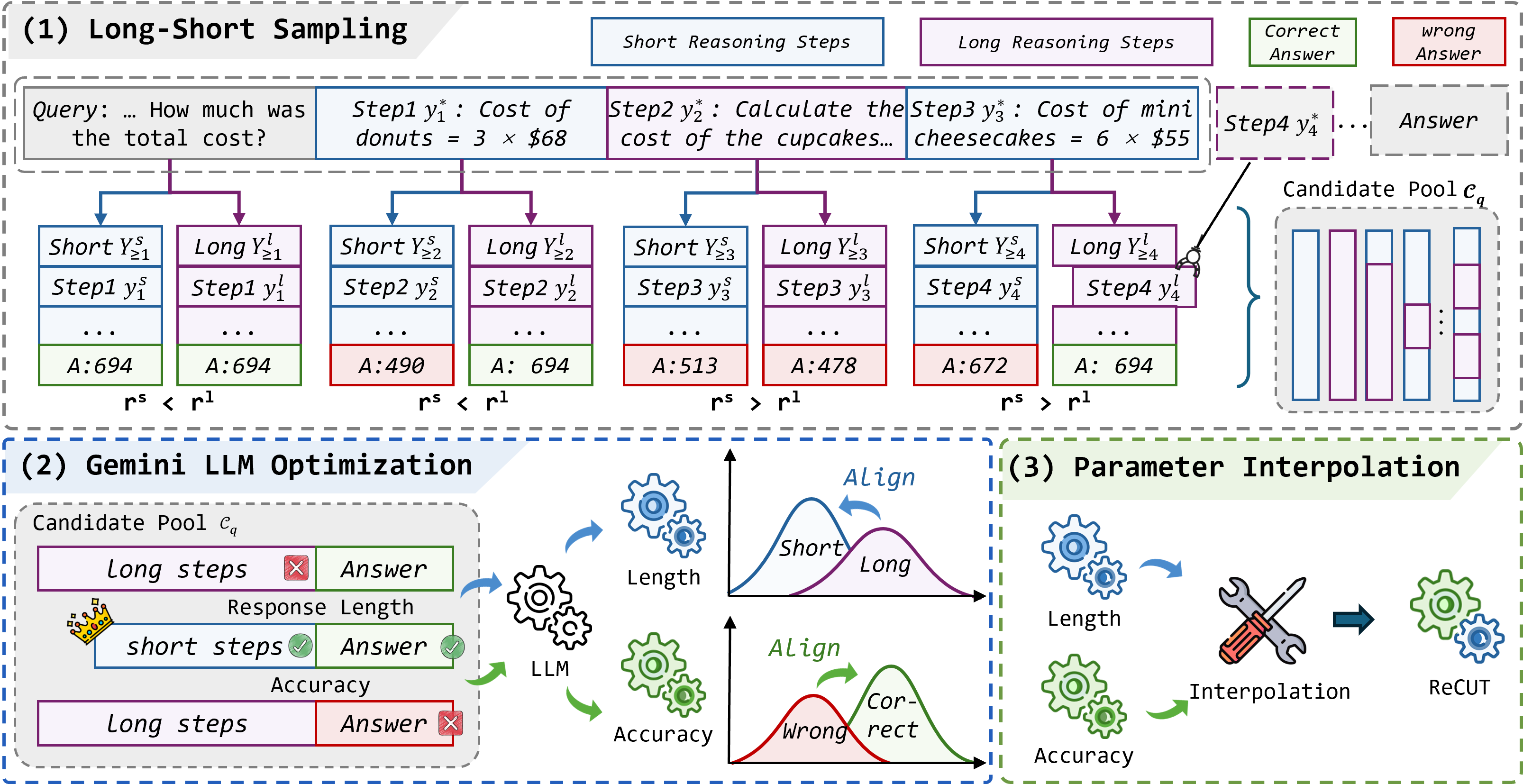}
    \caption{The Overview of Our Reasoning Compression Through Stepwise Trials (ReCUT) Model.}
    \label{fig:main}
\end{figure*}
This section presents Reasoning Compression Through Stepwise Trials (ReCUT), a method designed to reduce the reasoning length of Large Language Models (LLMs) while maintaining comparable performance. As shown in Figure~\ref{fig:main}, we first introduce the Long-Short Switch Sampling strategy, which adaptively constructs a reasoning trajectory pool $\mathcal{C}_q$ containing trajectories of varying lengths (Sec.~\ref{sec:3.1}). Then, we leverage these trajectories to train multiple Gemini-LLMs with complementary strengths and interpolate their parameters to balance reasoning accuracy and length (Sec.~\ref{sec:3.2}).

\subsection{Stepwise Reasoning Trajectory Exploration via Long-Short Sampling}\label{sec:3.1}
Given a question $q$, the reasoning trajectory produced by an LLM is denoted as $Y = \{y_{1}, \dots, y_\mathcal{T}, o\}$, where $y_{1:\mathcal{T}}$ are intermediate reasoning steps and $o$ is the final answer.

At step $t$, we design the \textit{Long-Short Switched Sampling} method to use both long and short prompting instructions to guide the LLM to generate the following reasoning trajectories $Y_{\ge t}$ with different lengths, based on the current optimal partial trajectory $Y_{<t}^* =\{y_1^*,...,y_{t-1}^*\}$. The sampled full trajectory $Y_{<t}^* \cup Y_{\ge t}$ is added to the trajectory pool $\mathcal{C}_q$. We then propose the \textit{Optimal Reasoning Step Selection} method. It designs a reward-based mechanism to select the optimal reasoning step  $y_t^*$, thereby updating the optimal trajectory $Y_{<t+1}^* =Y_{<t}^* \cup \{y_t^*\}$ fortrajectory sampling at the next step.

\textbf{Long-Short Switched Sampling.} At each step $t$, given the question $q$ and current optimal partial trajectory $Y^*_{<t}$, we prompt the LLM ($\mathcal{M}$) to generate two reasoning continuations--one long and one short--via different instructions:
\begin{equation}\label{eq:state}
\begin{aligned}
    Y^l_{\geq t} &= \mathcal{M}(\text{Instruct}_{l}(q,Y_{<t}^*)),\\
Y^s_{\geq t} &= \mathcal{M}(\text{Instruct}_{s}(q,Y_{<t}^*)),
\end{aligned}
\end{equation}
where $\text{Instruct}_{l}$ and $\text{Instruct}_{s}$ are tailored prompts that guide the model to produce reasoning of different lengths.
We then concatenate the existing partial trajectory with each of the new continuations to form full candidate trajectories:
\begin{equation}\label{eq:state}
Y^l_{[t]} =(Y^*_{<t} \cup Y^l_{\geq t}) , Y^s_{[t]} = (Y^*_{<t} \cup Y^s_{\geq t}),
\end{equation}
where the subscript $[t]$ indicates the concatenation occurs at step $t$. Both reasoning trajectories $Y^l_{[t]}$ and $Y^s_{[t]}$ are added to the candidate pool $\mathcal{C}_q$:
\begin{equation}
\mathcal{C}_q \leftarrow \mathcal{C}_q \cup \{Y^l_{[t]}\} \cup \{Y^s_{[t]}\}.
\end{equation}

\textbf{Optimal Stepwise Reasoning Selection.} Each candidate trajectory $Y_{[t]}$ is evaluated using a reward function that considers both accuracy and reasoning length:
\begin{equation}\label{eq:reward_mdp}
r(Y_{[t]}) =
\begin{cases}
\dfrac{1}{|Y_{[t]}|}, & \text{if } o = o^{\text{gold}}, \\[6pt]
-\dfrac{1}{|Y_{[t]}|},          & \text{if } o \neq o^{\text{gold}},
\end{cases}
\end{equation}
where $o^\text{gold}$ is the ground-truth answer and $|Y_{[t]}|$ denotes the number of tokens in $Y_{[t]}$. This reward function is designed to encourage the LLMs to explore trajectories that consider both length and accuracy. The reward function simultaneously encourages the correctness of the reasoning trajectory and penalizes its length. When the final answer in a trajectory is correct, we want the LLM to explore shorter trajectory directions to reduce the length. Conversely, when the final answer is incorrect, we prefer the LLM to explore longer trajectory directions, as incorrect answers may require a longer reasoning chain to be corrected.

We compare the rewards of the long and short trajectories and select the reasoning step $y^*_t$ from the trajectory with the higher reward:
\begin{equation}\label{eq:best}
y^*_t = \arg\max(r(Y^l_{[t]}), r(Y^s_{[t]})).
\end{equation}
The optimal trajectory is then updated by appending $y^*_t$:
\begin{equation}
Y^*_{<t+1} = Y^*_{<t} \cup {y^*_t}.
\end{equation}
This process is repeated iteratively until either the maximum step $\mathcal{T}$ is reached or the generation process terminates.

\subsection{Gemini LLM: Balancing Accuracy and Length via Parameter Interpolation}\label{sec:3.2}
In this section, we present Gemini LLM, a method that achieves a trade-off between the reasoning trajectory accuracy and length through parameter interpolation.

Given a question $q$ and its corresponding reasoning trajectory pool $\mathcal{C}_q$, we first construct two distinct preference datasets: $\mathcal{D}_{\text{acc}}$ and $\mathcal{D}_{\text{len}}$, each reflecting a different optimization objective. These datasets are used to train two specialized LLMs: $\mathcal{M}_\text{acc}$, which emphasizes answer accuracy, and $\mathcal{M}_{\text{len}}$, which encourages a shorter reasoning result.
The final Gemini model, $\mathcal{M}_{\text{merge}}$, is obtained via parameter interpolation between these two models, enabling it to balance the trade-off between accuracy and reasoning length during inference.

\textbf{Gemini LLM Optimization.} 
For each reasoning trajectory $Y_i \in \mathcal{C}_q$, we select the one with the correct final answer $o$ and the shortest length as the positive sample:
\begin{equation}\label{eq:best}
Y^+ = \arg\min_{Y_i \in \mathcal{C}_q}(|Y_i|)\; \text{s.t.} \; o = o^{\text{gold}},
\end{equation}
where $|Y_i|$ denotes the number of tokens in trajectory $Y_i$. 
% For each candidate reasoning trajectory $Y_i \in \mathcal{C}_q$, we compute its reward $r(Y_i)$ using the reward function $r(\cdot)$. The trajectory with the highest positive reward is selected as the positive sample $Y^+$:
% \begin{equation}\label{eq:best}
% Y^+ = \arg\max_{Y_i \in \mathcal{C}_q}(r(Y_i))\; \text{s.t.} \; r(Y_i) > 0 .
% \end{equation}
Next, for each $Y^+$, we select two negative samples from $\mathcal{C}_q$:
1) $Y^{-}_{\text{acc}}$ is the longest incorrect trajectory, i.e., a long trajectory that leads to an incorrect final answer;
2) $Y^{-}_{\text{len}}$ is the longest correct trajectory, i.e., a long trajectory that produces the correct answer. Formally, this can be described as:
% \begin{equation}\label{eq:best}
% \begin{aligned}
% Y^-_{\text{acc}} = \arg\min_{Y_i \in \mathcal{C}_q}(r(Y_i)) \; \text{s.t.} \; r(Y_i) < 0, \\
% Y^-_{\text{len}} = \arg\min_{Y_i \in \mathcal{C}_q}(r(Y_i)) \; \text{s.t.} \; r(Y_i) > 0.
% \end{aligned}
% \end{equation}
\begin{equation}\label{eq:best}
\begin{aligned}
Y^-_{\text{acc}} = \arg\max_{Y_i \in \mathcal{C}_q}(|Y_i|) \; \text{s.t.} \; o \neq  o^{\text{gold}}, \\
Y^-_{\text{len}} = \arg\max_{Y_i \in \mathcal{C}_q}(|Y_i|) \; \text{s.t.} \; o = o^{\text{gold}}.
\end{aligned}
\end{equation}
Using the identified samples, we construct two preference datasets: $\mathcal{D}_{\text{acc}}$ comprises triples $(q, Y^{+}, Y^{-}_{\text{acc}})$, which focuses on accuracy; $\mathcal{D}_{\text{len}}$ comprises triples $(q, Y^{+}, Y^{-}_{\text{len}})$, which focuses on length compression.

We then fine-tune the same base model $\mathcal{M}$ separately on these datasets using Direct Preference Optimization (DPO)~\cite{rafailov2023direct}:
\begin{equation}\label{eq:best}
\begin{aligned}
\mathcal{M}_{\text{len}} = \arg\min_{\mathcal{M}}\mathcal{L}_{\text{DPO}}(\mathcal{D}_{\text{len}}),\\
\mathcal{M}_{\text{acc}} = \arg\min_{\mathcal{M}}\mathcal{L}_{\text{DPO}}(\mathcal{D}_{\text{acc}}),
\end{aligned}
\end{equation}
where the DPO loss is defined as:
\begin{equation}\label{eq:dpo2}
\begin{aligned}
 & \mathcal{L}_{\text{DPO}}(\mathcal{D}) = -\mathbb{E}_{(q,Y^{+},Y^{-}) \sim \mathcal{D}} [\log \sigma (\\ &\beta \log \frac{\mathcal{M}(Y^{+} \mid q)}{\mathcal{M}^\text{ref}(Y^{+} \mid q)} - \beta \log \frac{\mathcal{M}(Y^{-} \mid q)}{\mathcal{M}^\text{ref} (Y^{-} \mid q)})],
\end{aligned}
\end{equation}
where $\beta$ is a hyperparameter and $\mathcal{M}^\text{ref}$ is a frozen reference model. Both $\mathcal{D}_{\text{acc}}$ and $\mathcal{D}_{\text{len}}$ share the same positive samples $Y^+$ but differ in their negative samples.

\textbf{Parameter Interpolation.} To combine the strengths of $\mathcal{M}_{\text{acc}}$ and $\mathcal{M}_{\text{len}}$, we perform parameter interpolation using the DARE-Ties strategy~\cite{yu2024language}:
\begin{equation}\label{eq:merge_parameters}
\mathcal{M}_{\text{merge}} = \theta_{\text{acc}} + \alpha \cdot \text{Top}_x(\theta_{\text{len}}),
\end{equation}
where $\theta_{\text{acc}}$ and $\theta_{\text{len}}$ are the parameters of $\mathcal{M}_{\text{acc}}$ and $\mathcal{M}_{\text{len}}$, respectively. Here, $\theta_{\text{acc}}$ serves as the base, and a sparsely selected fraction of parameters (controlled by $\text{Top}_x$) from $\theta_{\text{len}}$ are added with interpolation weight $\alpha$. This approach allows $\mathcal{M}_{\text{merge}}$ to incorporate the high-accuracy capability of $\mathcal{M}_{\text{acc}}$ and the reasoning compression capability of $\mathcal{M}_{\text{len}}$, effectively balancing performance and length.

\begin{table*}[t]
\centering
\small
\resizebox{\linewidth}{!}{
\begin{tabular}{l|rr|rr|rr|rr|rr|rr}
\hline
\multirow{2}{*}{\textbf{Model}} & \multicolumn{2}{c|}{\textbf{AIME24}} & \multicolumn{2}{c|}
{\textbf{AIME25}}& \multicolumn{2}{c|}
{\textbf{AMC23}} & \multicolumn{2}{c|}{\textbf{Math500}} & \multicolumn{2}{c|}{\textbf{GSM8K}} & \multicolumn{2}{c}{\textbf{Avg.}}\\ 
%\cline{2-15} 
~ &P@1 & \#Tok. & P@1 & \#Tok. & P@1& \#Tok.& P@1 & \#Tok.& P@1 & \#Tok. & P@1 & \#Tok. \\ \hline
%\rowcolor{gray!8} 

\multicolumn{13}{l}{\textbf{Qwen2.5-7B}} \\ \hline
Vanilla &\textbf{10.0} &3,252 &6.7 &2,855 &47.5 &2,512 &68.0 &1,570 &\textbf{87.4} &974 &43.9 &2,233 \\ 
CoD &0.0 &\textbf{419} &0.0 &\textbf{401} &15.0 &\textbf{162} &28.8 &\textbf{143} &41.7 &\textbf{71} &17.1 &\textbf{239}  \\ 
SFT &6.0 &4,055 &10.0 &3,159 &55.0 &1,992 &68.6 &1,560 &87.2 &899 &45.4 &2,333\\ 
Direct-DPO &\textbf{10.0}&3,887&3.3&2,407&\textbf{60.0}&2,496&\textbf{71.2}&1,625&87.3&972&\textbf{46.4}&2,277\\ 
L1 &3.3 &3,203 &\textbf{13.3} &3,191 &47.5 &3,108 &59.4 &3,053 &57.6 &3,366 &36.2 &3,184 \\
ReCUT &\textbf{10.0} &1,627 &\textbf{13.3} &1,670 &50.0 &1,425 &69.2 &1,062 &86.0 &704 &45.7 &1,298 \\ \hline
%\rowcolor{gray!8} 
\multicolumn{13}{l}{\textbf{Llama-3.1-8B}} \\ \hline
Vanilla &0.0 &7,943 &0.0 &6,188 &22.5 &7,518 &43.4 &3,718 &72.1 &1,290 &27.6&5,331\\ 
CoD &0.0 &\textbf{2,470} &0.0 &4,907 &20.0 &\textbf{1,851} &23.4 &1,720 &37.0 &\textbf{486} &16.1 &2,287\\ 
SFT&6.7 &10,802 &0.0 &6,332 &30.0 &5,013 &\textbf{45.8} &4,094 &\textbf{74.9} &1,246 &\textbf{31.5} &5,497\\ 
Direct-DPO &\textbf{10.0}&9,783&0.0&8,094&25.0&6,961&44.8&3,588&74.1&1,228&30.8&5,931\\ 
L1 &3.3 &4,705 &0.0 &6,185 &\textbf{32.5} &4,328 &44.2 &2,915 &61.0 &1,301 &28.2 &3,887 \\
ReCUT &6.7 &2,787 &0.0 &\textbf{3,182} &22.5 &1,879 &42.4 &\textbf{1,618} &73.9 &823 &29.1 &\textbf{2,058}\\ \hline
\end{tabular}}
\caption{Overall Performance. P@1 refers to the evaluation metric Pass@1, and \#Tok. indicates the number of tokens contained in the reasoning trajectory.}
\label{tab:main result}
\end{table*}

\section{Experimental Methodology}
This section first describes the datasets, evaluation metrics, and baselines, followed by the implementation details of our experiments.

\textbf{Dataset.} 
In our experiments, we follow prior work~\cite{aggarwal2025l1,li2025search} and adopt math reasoning datasets for training and evaluation of LLMs. Specifically, we randomly sample 8,000 math question-answer pairs from the DeepScaleR-Preview-Dataset~\cite{luo2025deepscaler} to construct our training set. This dataset comprises question-answer pairs collected from AIME, AMC, Omni-Math~\cite{gaoomni}, and STILL~\cite{min2024imitate}. For evaluation, we use math reasoning benchmarks spanning a range of difficulty levels, including GSM8K~\cite{cobbe2021training}, MATH500~\cite{hendrycks2measuring}, AMC23, AIME24 and AIME25. Except for GSM8K, all other evaluation datasets are sourced from the Math-AI repository\footnote{\url{https://huggingface.co/math-ai}}.

\textbf{Evaluation Metrics.} Following~\citet{li2025search}, we use Pass@1 to evaluate the final answer. And we also show the number of generated tokens (\#Token) to estimate the inference latency.

\textbf{Baselines.} We compare our ReCUT model against several baselines, including zero-shot prompting, Supervised Fine-Tuning (SFT), and Reinforcement Learning (RL) approaches.

We first consider two zero-shot baselines: Vanilla LLM and Chain of Draft (CoD)~\cite{xu2025chain}. For Vanilla LLM, we prompt the LLM to reason step-by-step to answer the question. CoD enhances this by introducing instructions that guide the LLM to generate concise and focused content at each reasoning step.
Next, for the SFT model, we select the reasoning trajectory with the highest reward from the candidate set $\mathcal{C}_q$ of the query $q$ and fine-tune the LLM to replicate it.
For the RL-based baselines, we consider Direct-DPO and L1~\cite{aggarwal2025l1}. Direct-DPO generates 20 reasoning trajectories per question and constructs a DPO training dataset by treating correct trajectories as positives and incorrect ones as negatives. L1 incorporates both reasoning length and answer accuracy into the reward function and trains the LLM using the GRPO algorithm~\cite{shao2024deepseekmath}.

% We use L1 as the baseline, which incorporates both reasoning length and accuracy into the reward, and trains the LLM using GPO.

% Method A designs specific prompts that encourage the LLM to generate concise and dense content at each step.

\textbf{Implementation Details.} In our experiments, we employ Qwen2.5-7B-Instruct~\cite{yang2024qwen2} and Llama-3.1-8B-Instruct~\cite{grattafiori2024llama} as backbones to implement all models. During dynamic step-wise sampling, we set different maximum exploration steps $\mathcal{T}$ for Qwen2.5-7B-Instruct and Llama-3.1-8B-Instruct, which are 8 and 12, respectively. During training, each model is trained for 1 epoch. We use LoRA~\cite{hulora} for efficient training. When using DARE-Ties\footnote{\url{https://github.com/arcee-ai/mergekit}} to implement parameter interpolation, we set the parameter density $\text{Top}_x$ and weight $\alpha$ to 0.25.

% evaluate all models on the test sets of four different reasoning datasets: AIME 2025, MATH (Hendrycks et al., 2021b), and AMC.
\section{Evaluation Results}
In this section, we first present the overall performance of ReCUT, followed by ablation studies to examine the contributions of its components. We then analyze the effectiveness of ReCUT in reasoning compression. Finally, we provide case studies for further illustration.

\begin{table*}[t]
\centering
\small
\resizebox{\linewidth}{!}{
\begin{tabular}{l|rr|rr|rr|rr|rr|rr}
\hline
\multirow{2}{*}{\textbf{Model}} & \multicolumn{2}{c|}{\textbf{AIME24}} & \multicolumn{2}{c|}{\textbf{AIME25}}& \multicolumn{2}{c|}{\textbf{AMC23}} & \multicolumn{2}{c|}{\textbf{Math500}} & \multicolumn{2}{c|}{\textbf{GSM8K}} & \multicolumn{2}{c}{\textbf{Avg.}}\\ 
%\cline{2-15} 
~ &P@1 & \#Tok.& P@1 & \#Tok. & P@1& \#Tok.& P@1 & \#Tok.& P@1 & \#Tok. & P@1 & \#Tok. \\ \hline
%\rowcolor{gray!8} 
\multicolumn{13}{l}{\textbf{Qwen2.5-7B}} \\ \hline
ReCUT &10.0 &\textbf{1,627} &\textbf{13.3} &1,670 &50.0 &\textbf{1,425}&69.2 &\textbf{1,062} &86.0 &\textbf{704} &45.7 &\textbf{1,298}\\ 
w/o Explore&13.3 &3,073 &3.3 &2,275 &45.0 &1,945 &67.8 &1,401 &80.3 &811 &41.9 &1,901\\ 
w/o Prompt &3.3 &1,753 &6.7 &1,808 &50.0 &1,528 &63.3 &1,183 &80.3 &775 &40.7 &1,409\\
Only $\mathcal{M}_{\text{acc}}$&\textbf{17.0}&2,231&3.3&2,232&\textbf{57.5}&2,464&\textbf{70.0} &1,491 &86.6&910&\textbf{46.9}&1,866  \\ 
Only $\mathcal{M}_{\text{len}}$&10.0&1,682&\textbf{13.3}&\textbf{1,640}&50.0&1,815&68.0 &1,102 &\textbf{87.4}&703&45.7 &1,388 \\ \hline
%\rowcolor{gray!8} 
\multicolumn{13}{l}{\textbf{Llama-3.1-8B}} \\ \hline
ReCUT &\textbf{6.7} &2,787 &0.0 &3,182 &\textbf{22.5} &1,879 &42.4 &1,618 &73.9 &823 &\textbf{29.1} &2,058\\ 
w/o Explore &3.3 &7,389 &0.0 &5,880 &17.5 &6,806 &42.0 &3,199 &64.8 &1,224 &25.2 &4,900\\ 
w/o Prompt &\textbf{6.7} &2,458 &0.0 &\textbf{2,122} &\textbf{22.5} &1,887 &39.6 &1,524 &74.4 &\textbf{797} &28.6 &\textbf{1,758}\\
Only $\mathcal{M}_{\text{acc}}$ &3.3 &2,785 &0.0 &4,801 &\textbf{22.5} &2,842 &\textbf{43.4} &1,954 &\textbf{74.8} &887 &28.8 &2,654  \\ 
Only $\mathcal{M}_{\text{len}}$ &0.0 &\textbf{2,313} &0.0 &2,293 &17.5 &\textbf{1,859} &41.6 &\textbf{1,562} &73.9 &817 &26.6 &1,769 \\ 
\hline
\end{tabular}}

\caption{Ablation Study. Both ReCUT (Only $\mathcal{M}_{\text{acc}}$) and ReCUT (Only $\mathcal{M}_{\text{len}}$) are Gemini LLMs.}
\label{tab:ablation}
\end{table*}
\subsection{Overall Performance}
Table~\ref{tab:main result} presents the overall performance of ReCUT and baseline methods across different mathematical tasks and backbone models.

ReCUT consistently achieves the shortest reasoning trajectories across these math tasks of varying difficulty, while maintaining comparable accuracy to baseline models. This demonstrates the effectiveness of ReCUT in achieving an optimal trade-off between reasoning accuracy and efficiency. Notably, ReCUT surpasses the Vanilla LLM on most tasks, achieving higher Pass@1 scores while consuming only about half the number of tokens. This indicates that ReCUT significantly reduces inference cost and enables more efficient utilization of computational resources. Moreover, ReCUT demonstrates strong generalization ability, consistently improving both accuracy and reasoning compression over Vanilla LLMs across different foundation models, such as Qwen and Llama.

Among baseline methods, CoD effectively shortens reasoning trajectories but suffers a notable drop in accuracy, suggesting that prompting LLMs to compress reasoning without proper guidance may lead to the omission of critical information. In contrast, both SFT and Direct-DPO improve accuracy by fine-tuning LLMs with the shortest correct reasoning paths but fail to compress the reasoning length effectively. ReCUT leverages preference-based learning to better guide LLMs in balancing accuracy and length. Furthermore, ReCUT outperforms the L1 model trained via GRPO, which also considers both accuracy and length rewards. ReCUT achieves significantly higher Pass@1 scores while substantially reducing token usage, demonstrating its ability to effectively integrate signals from both reasoning quality and efficiency.

%On Qwen2.5-7B, ReCUT outperforms the Vanilla LLM on the majority of tasks. For instance, in the AIME24 task, ReCUT achieves a Pass@1 of 10.0\%, surpassing the Vanilla LLM by 6.0\%. Additionally, it consumes only 1,627 tokens---approximately half the number used by the Vanilla LLM---indicating a much more efficient use of computational resources. A similar pattern is observed in tasks such as AIME25 and AMC23: while the SFT model improves the Pass@1 score, it also leads to increased token usage. Furthermore, ReCUT outperforms the L1 baseline trained via GRPO, achieving higher Pass@1 while significantly reducing token consumption. Thus, ReCUT exhibits its effectiveness in keeping both efficiency and accuracy during reasoning.
%As demonstrated by the experimental results on Llama-3.1-8B, ReCUT significantly reduces token usage while maintaining the same Pass@1 performance. On the AIME24 benchmark, ReCUT achieves a Pass@1 rate of 6.7\%, substantially outperforming both the Vanilla LLM and the SFT model, which achieve 0.0\%. Although ReCUT incurs slightly higher token consumption than the SFT model, it delivers markedly better performance and still requires far fewer tokens than the standard method. Furthermore, on GSM8K, ReCUT improves the Pass@1 rate by 2\% while reducing token usage by 40\%, highlighting its efficiency even on simpler tasks.

\subsection{Ablation study}

In this section, we present ablation studies to evaluate the effectiveness of different components in our ReCUT framework.

As shown in Table~\ref{tab:ablation}, we compare ReCUT with four ablated variants: ReCUT w/o Explore, ReCUT w/o Prompt, ReCUT (Only $\mathcal{M}_{\text{acc}}$), and ReCUT (Only $\mathcal{M}_{\text{len}}$). Specifically, ReCUT w/o Explore replaces the stepwise sampling strategy with direct long-short sampling for generating reasoning trajectories. ReCUT w/o Prompt retains stepwise sampling but removes the long-short switch prompt, using a unified prompt that does not control reasoning length during sampling. Both ReCUT (Only $\mathcal{M}_{\text{acc}}$) and ReCUT (Only $\mathcal{M}_{\text{len}}$) are LLMs trained to conduct an accurate answer and encourage a shorter reasoning result, respectively. 

Compared to the full ReCUT model, both ReCUT w/o Explore and ReCUT w/o Prompt exhibit a clear drop in reasoning accuracy, demonstrating the effectiveness of both stepwise sampling and the long-short instruction mechanisms. The performance gap between these two variants further highlights that stepwise sampling contributes more significantly to improving reasoning compression during DPO training. Both ReCUT (Only $\mathcal{M}_{\text{acc}}$) and ReCUT (Only $\mathcal{M}_{\text{len}}$) achieve comparable overall performance but exhibit distinct preferences: the former tends to generate longer reasoning results to maximize accuracy, while the latter sacrifices some accuracy to produce shorter reasoning outputs. By interpolating the parameters of these two models, ReCUT can better balance the reasoning accuracy and length. More concretely, ReCUT reduces reasoning length while maintaining accuracy when using Qwen2.5-7B as the backbone. With Llama-3.1-8B, ReCUT improves accuracy at the cost of generating slightly longer reasoning results.

\begin{figure}[t]
  \centering
  % 第一个子图
  \subfigure[Performance of the Qwen2.5-7B based Models.]{
    \includegraphics[width=0.45\linewidth]{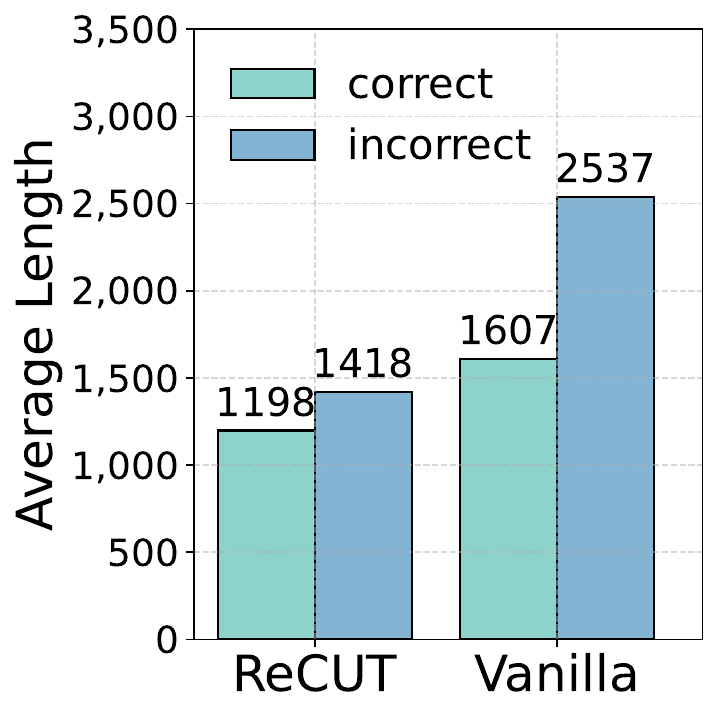}
    \label{fig:average length:qwen}
  }
  % 第二个子图
  \subfigure[Performance of the Llama-3.1-8B based Models.]{
    \includegraphics[width=0.45\linewidth]{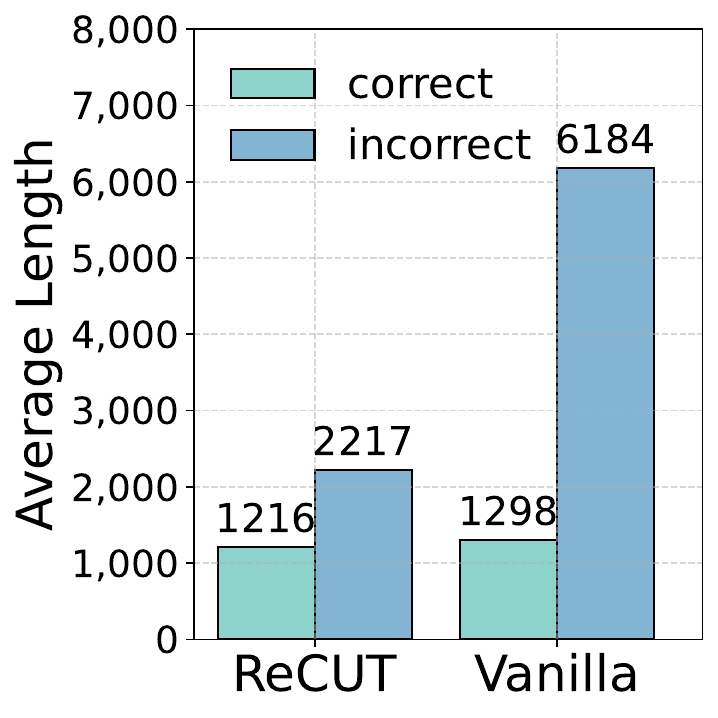}
    \label{fig:average length:llama3}
  }
  \caption{Average Lengths of Reasoning Outputs Across Models. We plot the output lengths of ReCUT and Vanilla LLMs in scenarios where the models produce correct and incorrect answers. All evaluation datasets are used in this experiment.}
  \label{fig:average}
\end{figure}
\begin{figure}[t]
  \centering
  % 第一个子图
  \subfigure[Reasoning Steps Required for Solving GSM8K Questions.]{
    \includegraphics[width=0.46\linewidth]{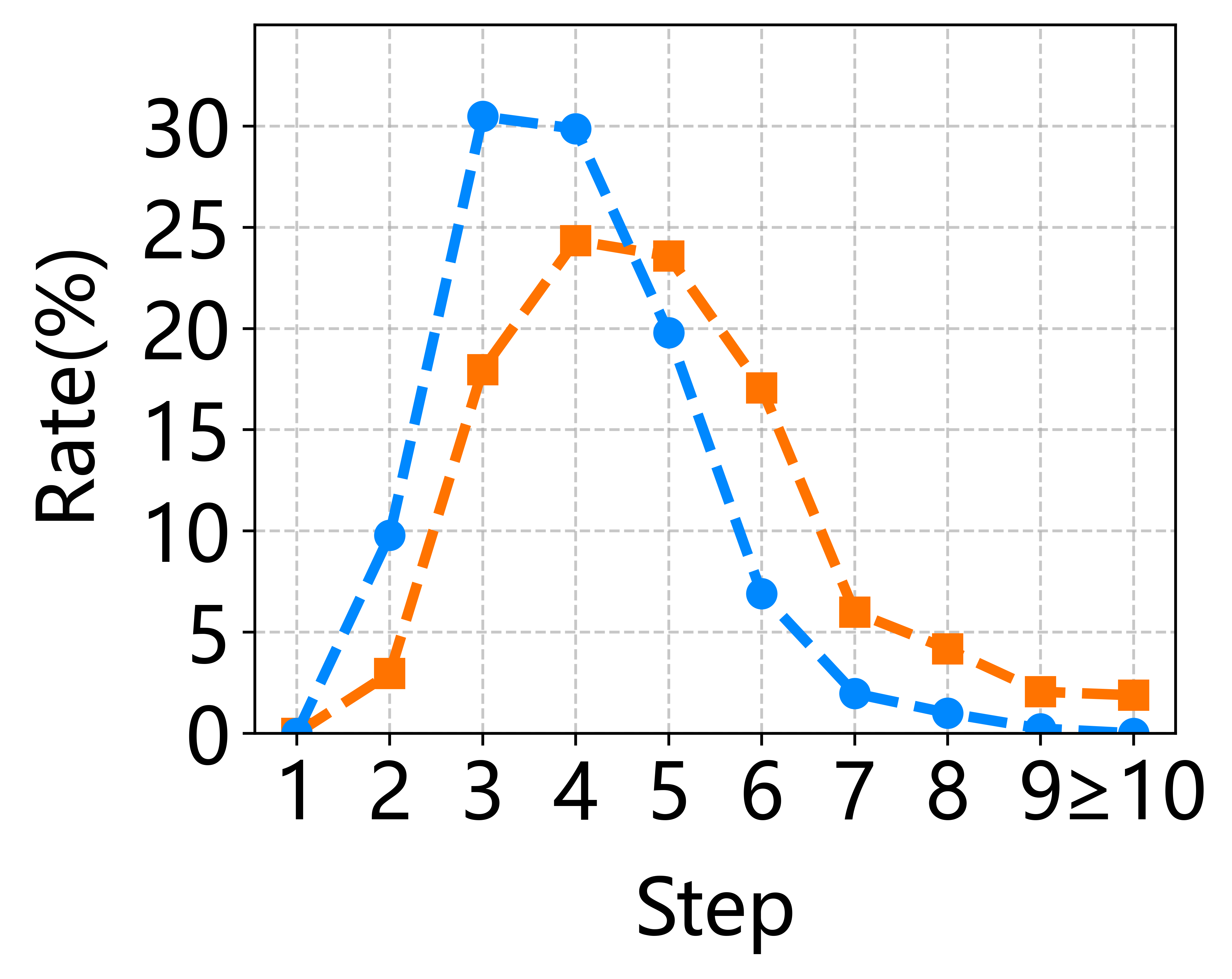}
    \label{fig:reasoning_qwen:step_rate_gsm8k}
  }
  \subfigure[Reasoning Steps Required for Solving AMC23 Questions.]{
    \includegraphics[width=0.46\linewidth]{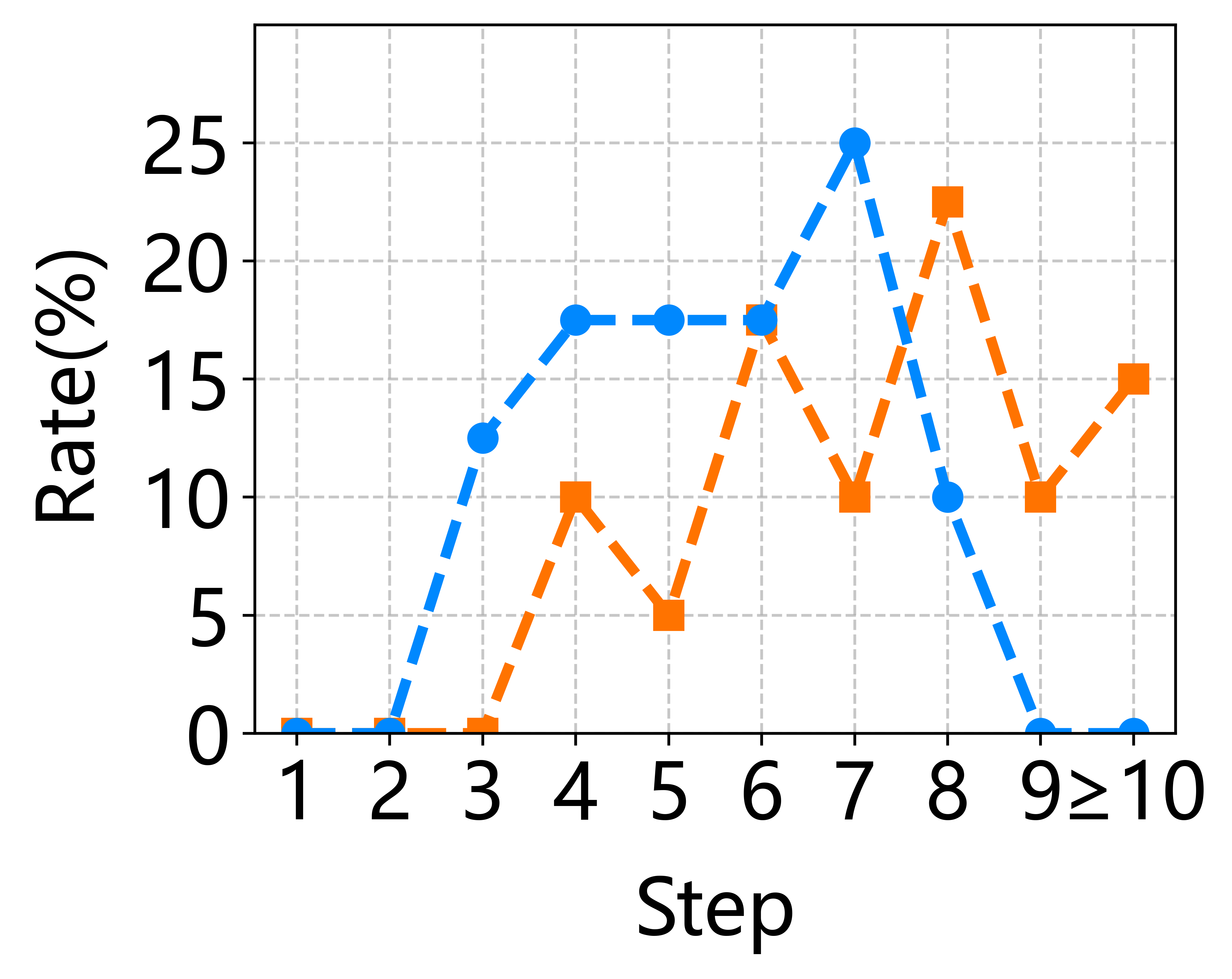}
    \label{fig:reasoning_qwen:step_rate_amc}
  }
   \subfigure[Accuracy vs. Reasoning Steps on GSM8K.] { \label{fig:reasoning_qwen:step_acc_gsm8k} 
    \includegraphics[width=0.48\linewidth]{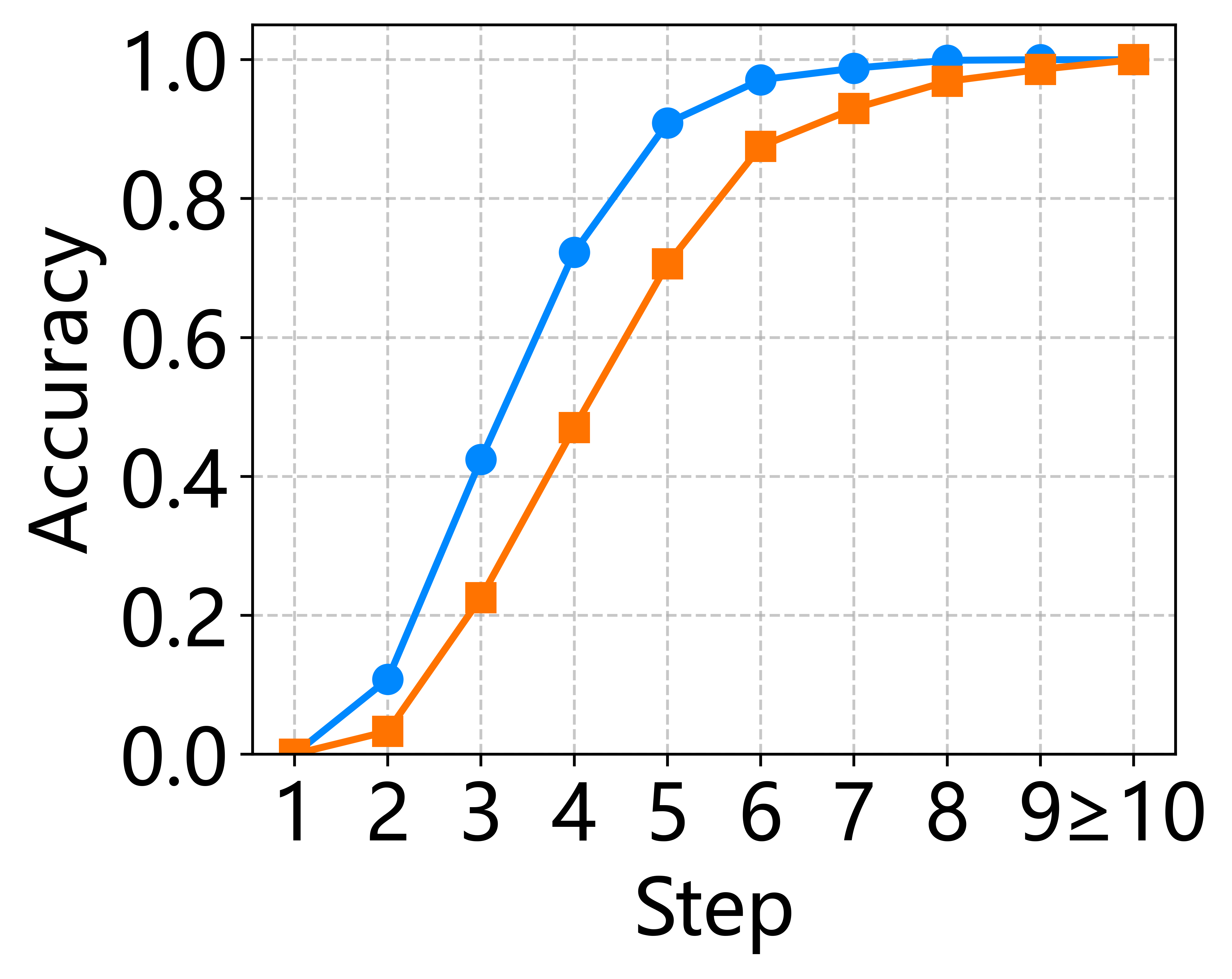}}
      \subfigure[Accuracy vs. Reasoning Steps on AMC23.] { \label{fig:reasoning_qwen:step_acc_amc} 
    \includegraphics[width=0.48\linewidth]{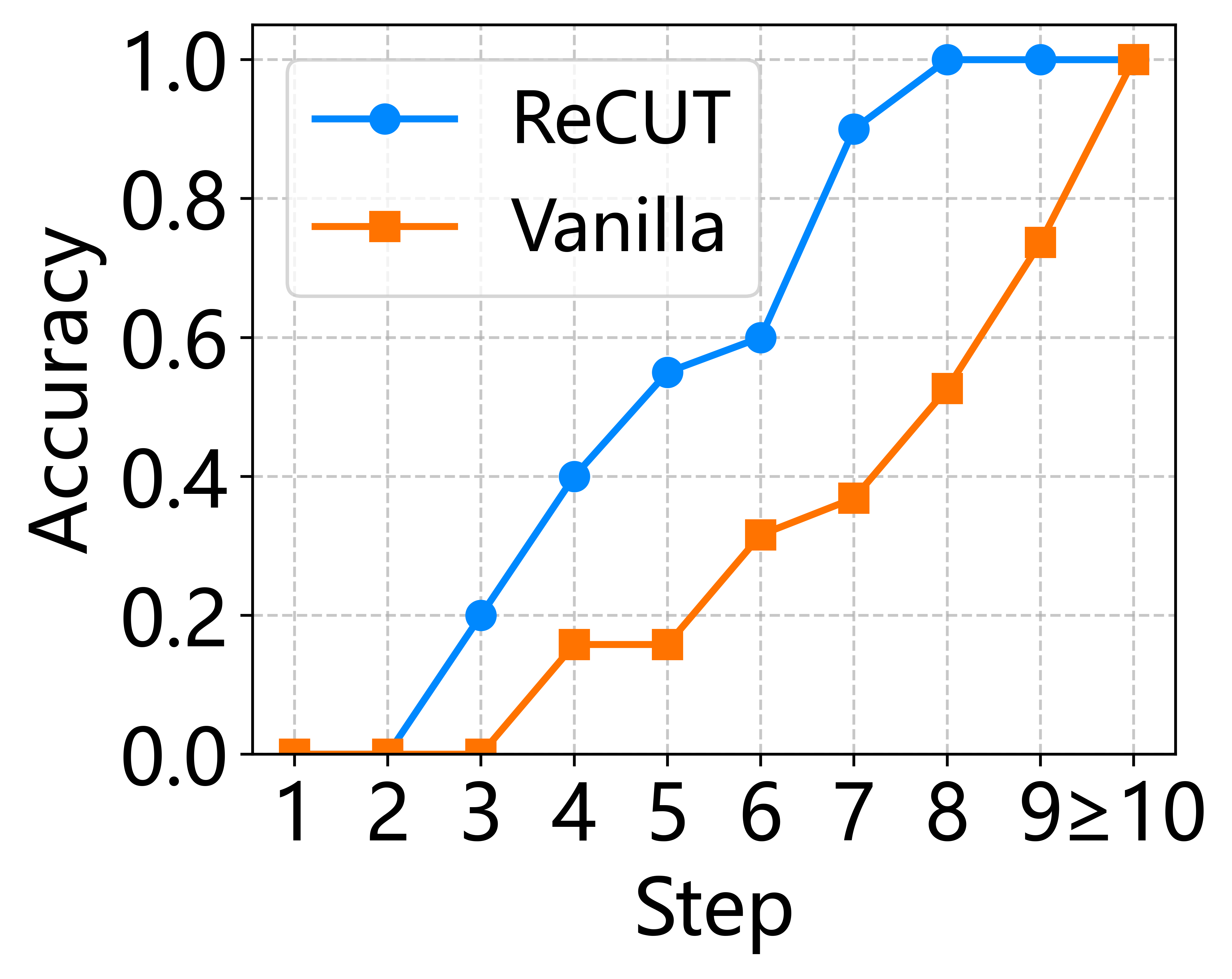}}
    
  \caption{Performance of Different Models Across Reasoning Steps.
Both the Vanilla LLM and ReCUT variants are implemented using Qwen2.5-7B. All models are evaluated on GSM8K and AMC23.}
  \label{fig:reasoning_qwen}
\end{figure}
\subsection{The Effectiveness of ReCUT in Reasoning Compression}
This experiment evaluates the effectiveness of ReCUT in compressing the reasoning outcomes. We begin by presenting the lengths of the generated reasoning trajectories, followed by an analysis of the reasoning steps used to solve problems.

\textbf{Average Length.} We first generate reasoning outputs using both the Vanilla LLM and ReCUT across all evaluation datasets. To better understand the behavior of each model, we categorize the outputs into two groups: correct and incorrect, depending on whether the final answer is accurate.

As shown in Figure~\ref{fig:average}, for the Vanilla LLM, incorrect reasoning trajectories tend to be longer than correct ones, particularly for Llama-3.1-8B. This suggests that when faced with challenging problems, LLMs are prone to overthinking---producing unnecessarily lengthy and complex reasoning chains that consume more computational resources while ultimately leading to incorrect answers. In contrast, ReCUT significantly shortens both correct and incorrect reasoning trajectories and reduces the gap in length between them. This demonstrates the ability of ReCUT to compress reasoning effectively and eliminate redundant content, even under difficult reasoning scenarios.
\begin{figure*}[t] 
\centering
    \includegraphics[width=1.0\textwidth]{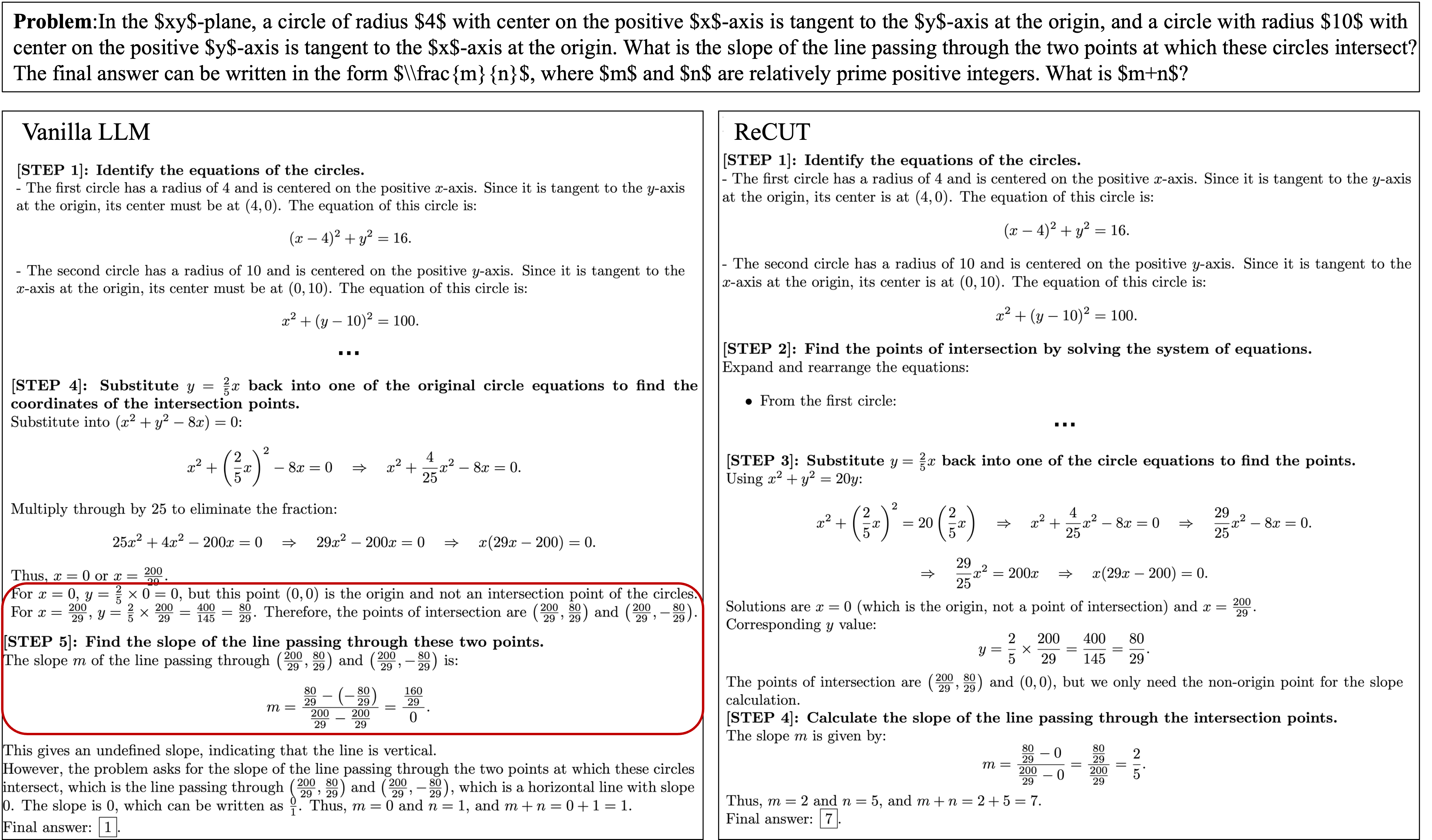}
    \caption{Case Study. A comparison between the Vanilla LLM and ReCUT (Qwen2.5-7B). The red box marks the first step in the reasoning process where the model begins to exhibit errors.} \label{fig:case}
\end{figure*}

\textbf{Reasoning Performance Across Steps.} As shown in Figure~\ref{fig:reasoning_qwen}, we collect the reasoning trajectories generated by ReCUT and the Vanilla LLM on the AMC23 and GSM8K datasets. We retain only those trajectories that lead to correct answers and analyze their distribution across reasoning steps. All models are based on Qwen2.5-7B. Results for models implemented with Llama-3.1-8B are provided in Appendix~\ref{app:reasoning_llama}.

Figures~\ref{fig:reasoning_qwen:step_rate_gsm8k} and \ref{fig:reasoning_qwen:step_rate_amc} illustrate the number of reasoning steps required to solve problems. ReCUT consistently reduces the number of reasoning steps across all evaluation settings. On the relatively simpler GSM8K dataset, ReCUT solves most problems within approximately 4 steps, outperforming the Vanilla LLM. On the more complex dataset AMC23, ReCUT typically solves problems in fewer than 10 steps, indicating its ability to construct more efficient reasoning trajectories by incorporating necessary information early on and avoiding unnecessarily lengthy chains of reasoning.

Figures~\ref{fig:reasoning_qwen:step_acc_gsm8k} and \ref{fig:reasoning_qwen:step_acc_amc} present the reasoning accuracy at each step. ReCUT consistently reaches correct answers in fewer steps compared to the Vanilla LLM. On GSM8K, both ReCUT and the Vanilla LLM generate correct reasoning trajectories in over 80\% of cases within 6 steps. However, the advantage of ReCUT becomes more evident on the challenging AMC23 dataset, where it achieves 60\% accuracy within just 7 steps---significantly outperforming the Vanilla LLM. These results demonstrate that ReCUT effectively shortens reasoning trajectories without compromising accuracy, mitigating the overthinking problem commonly observed in Vanilla LLMs.

\subsection{Case Study}
In this section, we randomly select a case from AMC23 to demonstrate the effectiveness of ReCUT in balancing the accuracy and length of reasoning trajectories. In this case, the question is a math competition problem that is relatively difficult and contains traps in the problem-solving process.

As shown in Table~\ref{fig:case}, the overall number of steps in the reasoning trajectory generated by Vanilla LLM is significantly greater than that of ReCUT. To derive the intermediate variable ``\( x = 0 \) or \( x = \frac{200}{29} \)'', Vanilla LLM takes nearly four reasoning steps, whereas ReCUT accurately reaches this result midway through its third reasoning step. This indicates that ReCUT significantly improves the reasoning efficiency of the LLM.
In the subsequent calculation process, Vanilla LLM produces incorrect intersection points of the circles, which leads to a division by zero when computing the slope. Although Vanilla LLM attempts self-reflection to resolve the issue, it still outputs an incorrect answer. In contrast, ReCUT accurately identifies the intersection points and completes the slope calculation using a much more concise reasoning trajectory, arriving at the correct answer. This demonstrates that ReCUT effectively balances the reasoning length and accuracy by producing correct answers with fewer reasoning steps.

\section{Conclusion}
In this paper, we introduce ReCUT (Reasoning Compression Through Stepwise Trials), an effective method designed to optimize LLMs to mitigate the overthinking issue. ReCUT introduces a stepwise reasoning trajectory
exploration mechanism to construct a more diverse reinforcement learning training dataset and train Gemini LLMs to balance accuracy and length via parameter interpolation. Our experimental analysis reveals that ReCUT achieves considerable improvements in token efficiency and maintains or enhances accuracy across different difficulty levels of math tasks and various backbone models. 

% Through the innovative use of stepwise long-short switched sampling combined with a carefully designed reward-based preference optimization strategy, ReCUT successfully reduces reasoning trajectory lengths without sacrificing accuracy. Our experimental analysis reveals that ReCUT achieves considerable improvements in token efficiency and maintains or enhances accuracy across different difficulty levels of math tasks and various backbone models. 
\section*{Limitation}
Although Model ReCUT demonstrates effectiveness in balancing the accuracy and length of the generated reasoning trajectories, there are still some limitations. First, the stepwise seasoning trajectory exploration strategy of the ReCUT is limited by the instruction-following capability of the LLMs. If the LLMs fail to effectively follow the designed instructions to stepwise generate reasoning trajectories, it will affect the quality of the constructed preference dataset. Furthermore, the parameter selection in the parameter interpolation method, DARE-Ties, is based on empirical practices from prior work. When merging models with relatively weaker performance, they typically set the weight and density around 0.3-0.4 to prevent degrading the performance of the merged model. Therefore, we don't conduct further analysis or experiments on parameter selection.
\bibliography{custom}
\clearpage
\newpage
\appendix

\section{Appendix}

\subsection{License}
We show the licenses for our use of the datasets, AIME24, AMC23 are not currently labeled with license types,  MATH500 is licensed under the Apache License 2.0, AIME25 and GSM8K is licensed under the MIT license.
\begin{figure}[t]
    \centering

  % 第四个子图
  \subfigure[Reasoning Steps Required
for Solving GSM8K Questions.]{
    \includegraphics[width=0.46\linewidth]{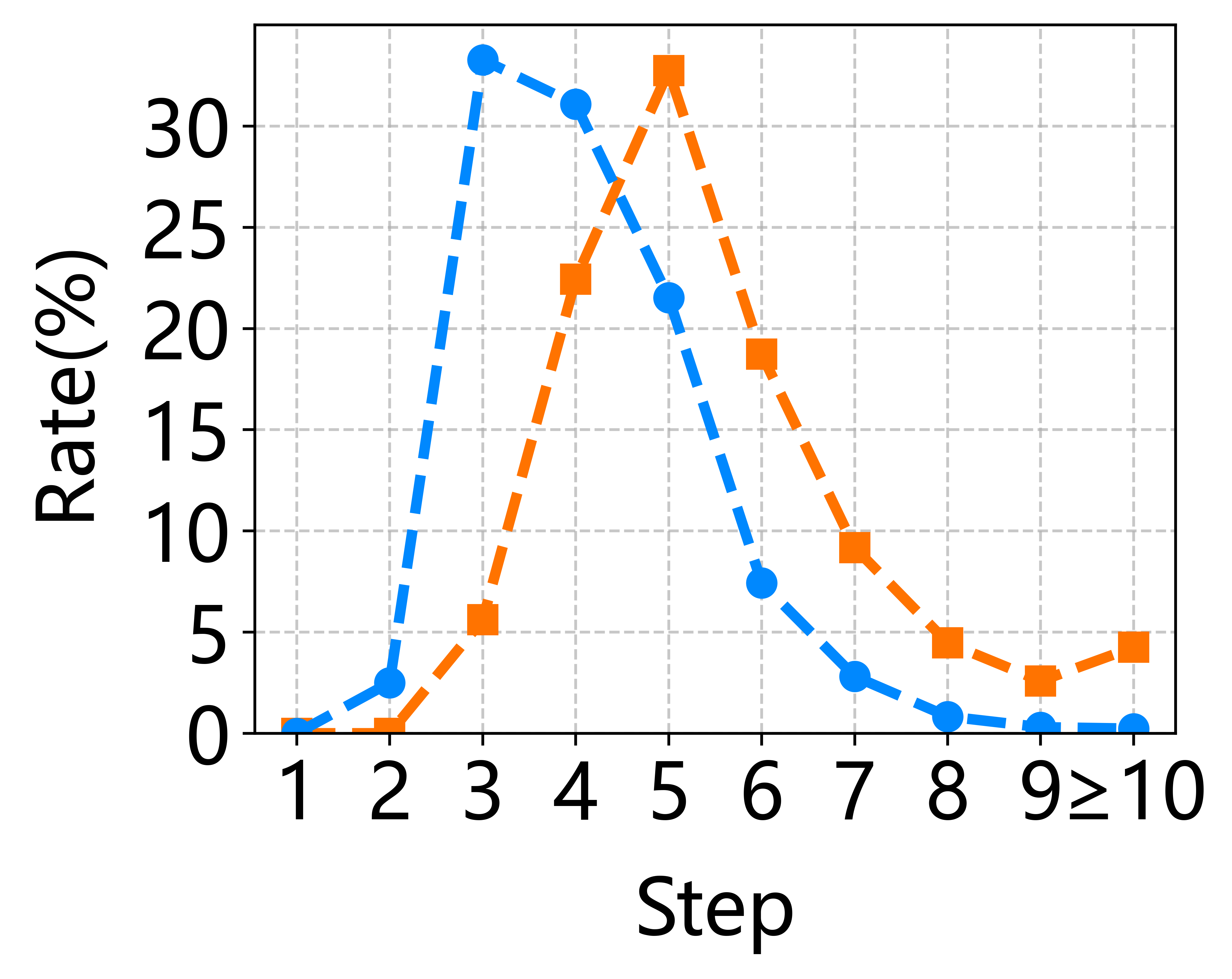}
    \label{fig:reasoning_llama:step_rate_gsm8k}
  }
    \subfigure[Reasoning Steps Required
for Solving AMC23 Questions.]{
    \includegraphics[width=0.46\linewidth]{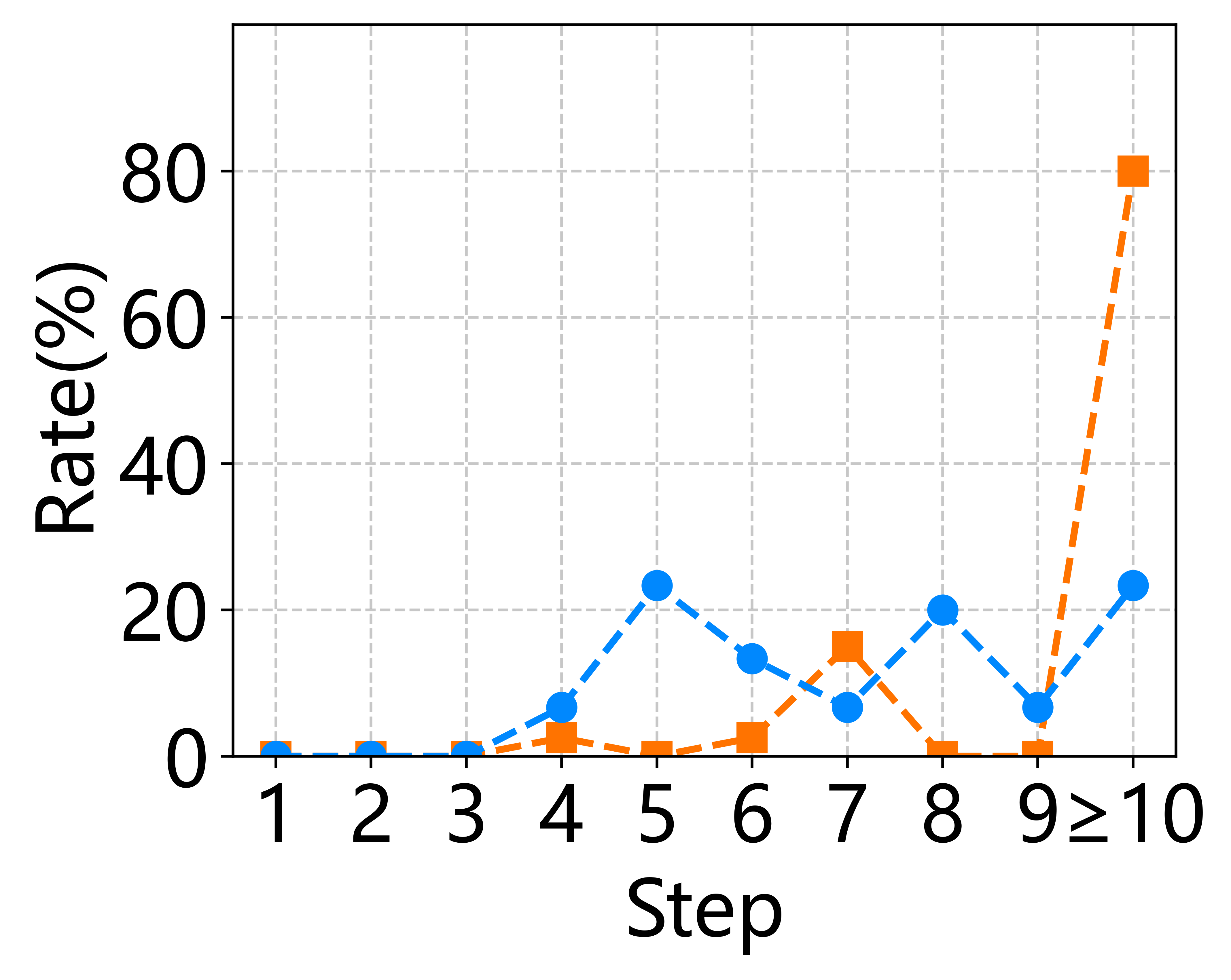}
    \label{fig:reasoning_llama:step_rate_amc}
  }
    \subfigure[Accuracy vs. Reasoning
Steps on GSM8K.] { \label{fig:reasoning_llama:step_acc_gsm8k}
    \includegraphics[width=0.48\linewidth]{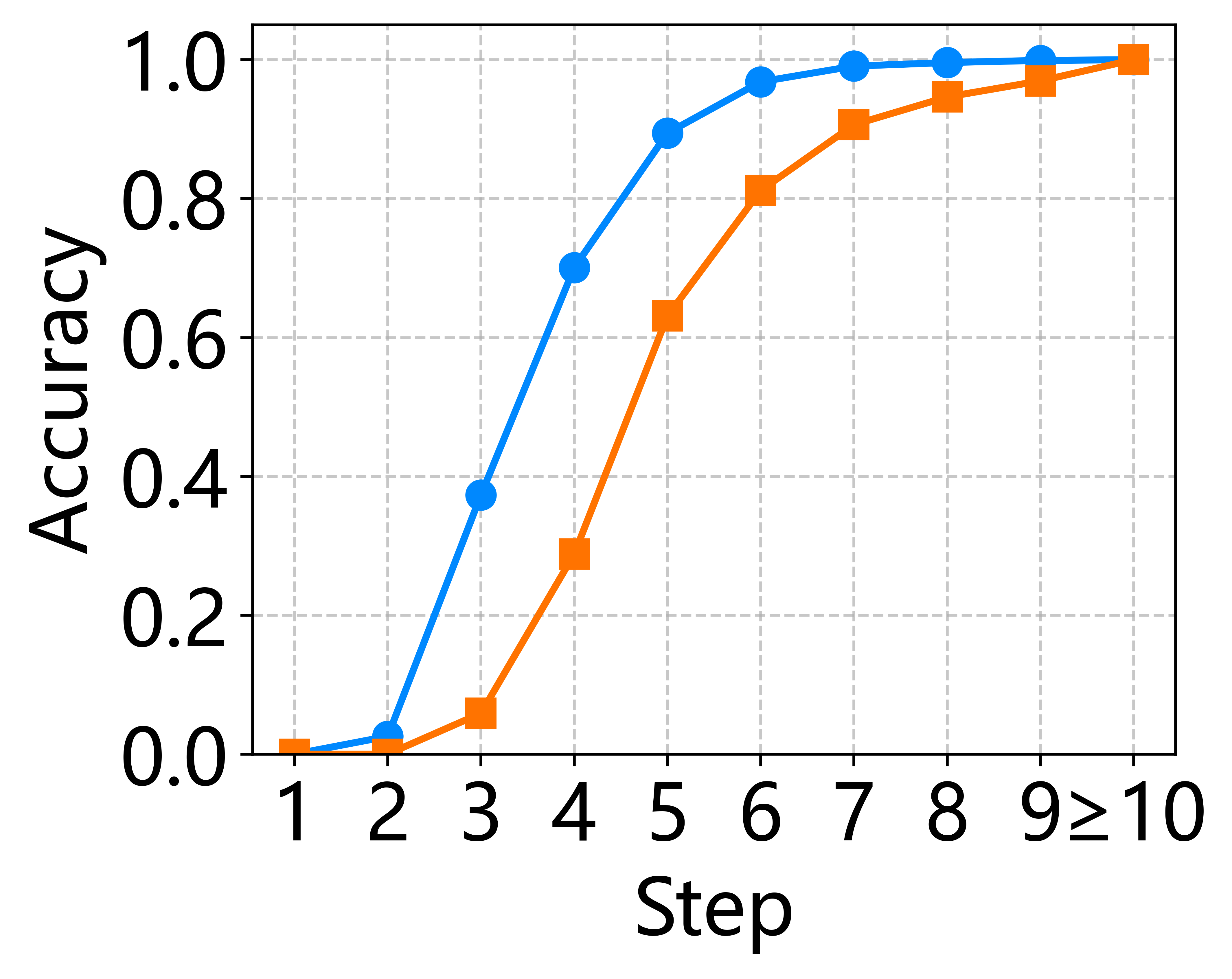}}
    \subfigure[Accuracy vs. Reasoning
Steps on AMC23.] { \label{fig:reasoning_llama:step_acc_amc} 
    \includegraphics[width=0.48\linewidth]{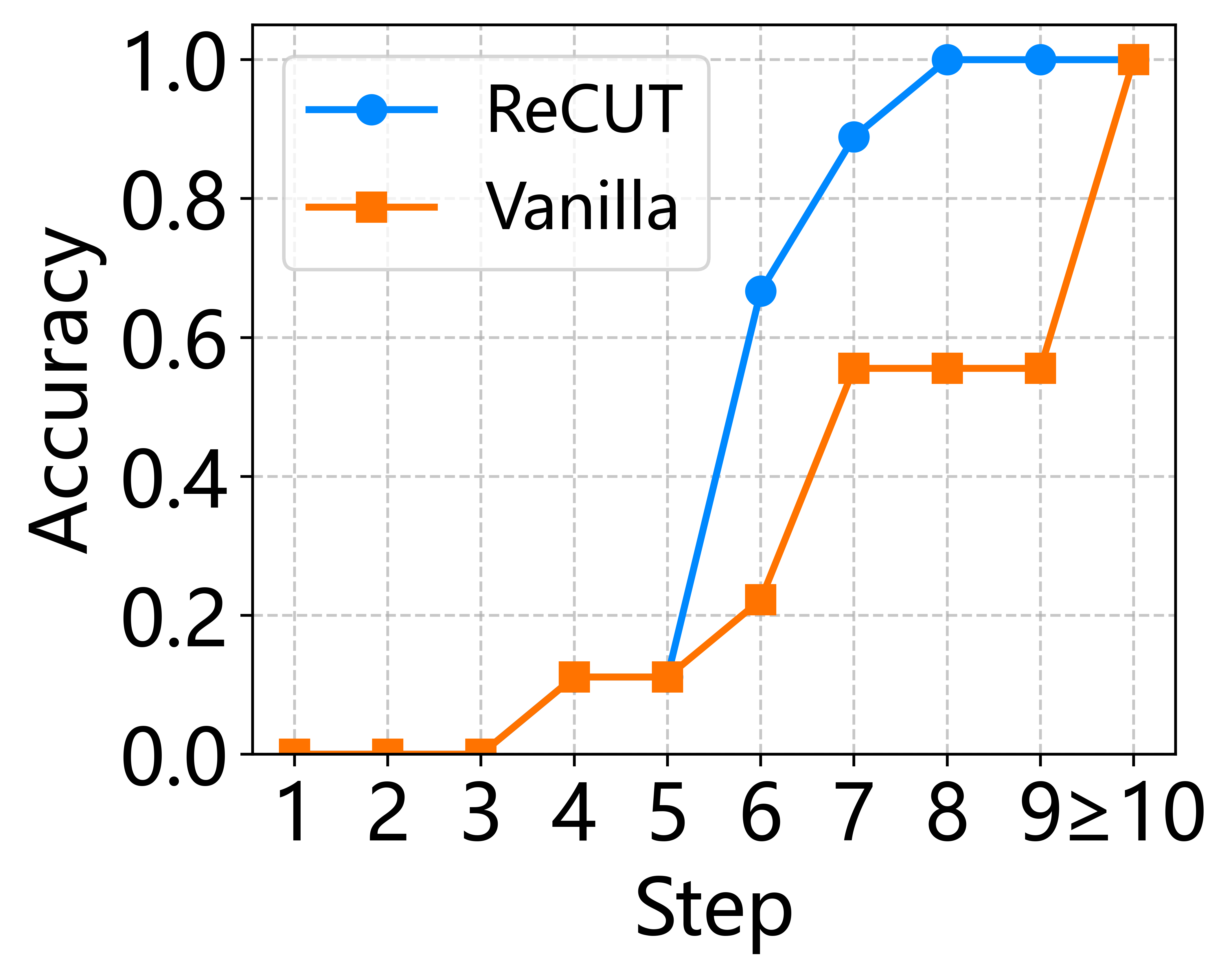}}
    \caption{Performance of Different Models Across Reasoning Steps.
Both the Vanilla LLM and ReCUT variants are implemented using Llama3.1-8B. All models are evaluated on GSM8K and AMC23.}
    \label{fig:reasoning_llama}
\end{figure}

\subsection{Reasoning Performance of Llama-Based Models}\label{app:reasoning_llama}
We conduct additional experiments, as shown in Figure~\ref{fig:reasoning_llama}, to further evaluate the effectiveness of Llama-based models. Specifically, we compare the performance of the Vanilla LLM and the ReCUT model on two benchmarks: GSM8K and AMC23.

As illustrated in Figures~\ref{fig:reasoning_llama:step_rate_gsm8k} and \ref{fig:reasoning_llama:step_rate_amc}, ReCUT consistently requires fewer reasoning steps to arrive at correct answers, further validating its effectiveness in reasoning compression. Notably, for the more challenging AMC23 dataset, the Vanilla LLM requires significantly more reasoning steps than Qwen2.5-7B (Figure~\ref{fig:reasoning_qwen:step_rate_amc}), which can be attributed to differences in backbone model capability. Even under such conditions, ReCUT substantially reduces the number of reasoning steps required, highlighting its strong generalization ability across different model backbones.
In addition, Figures~\ref{fig:reasoning_llama:step_acc_gsm8k} and \ref{fig:reasoning_llama:step_acc_amc} show the step-wise reasoning accuracy of different models. ReCUT consistently achieves higher accuracy with fewer steps, demonstrating its ability to generate more efficient and effective reasoning trajectories.

\begin{figure}[t] 
\centering
    \includegraphics[width=0.5\textwidth]{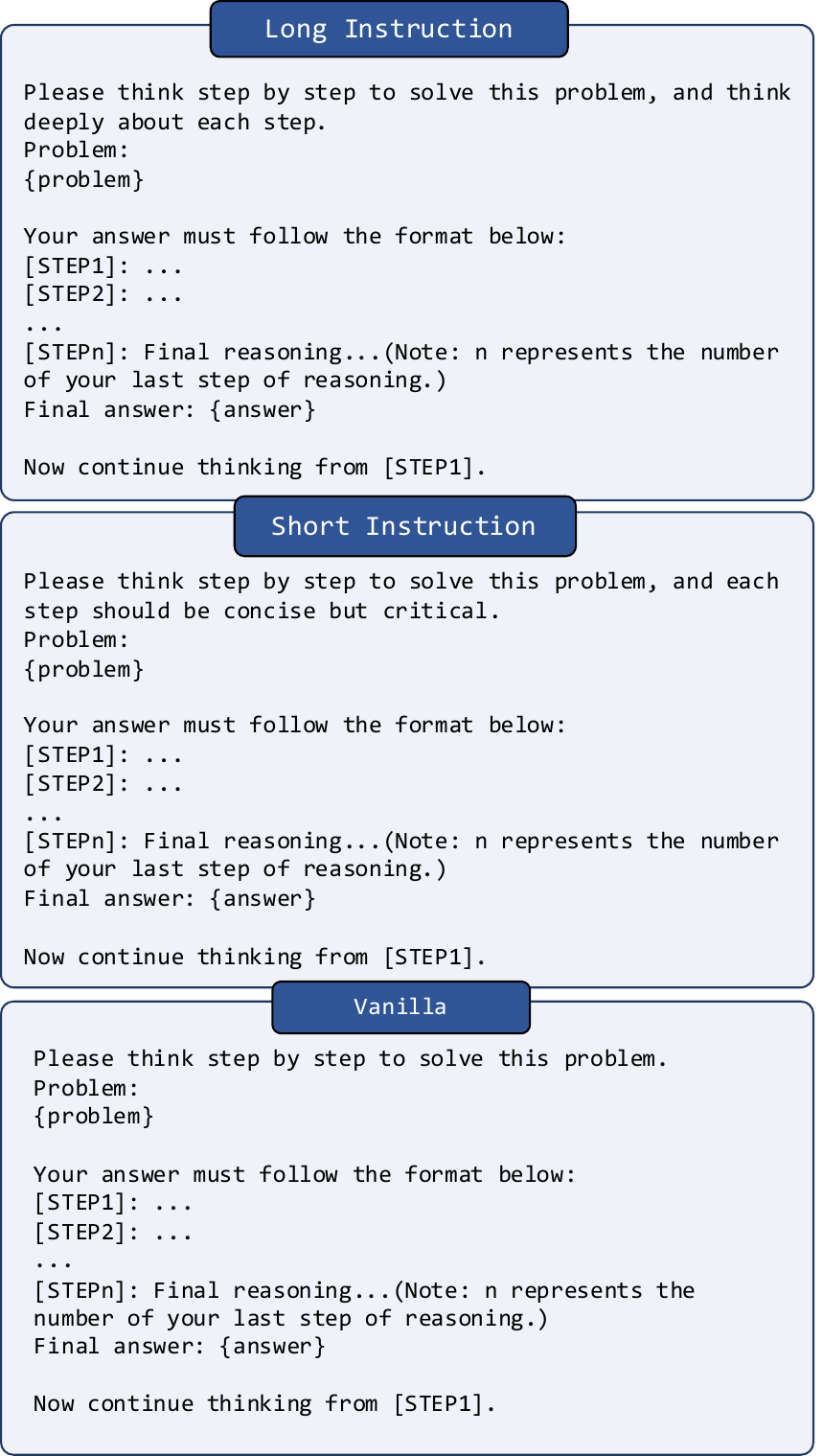}
    \caption{The Instruction Used in Our Experiments.} \label{fig:longtoshort_prompt}
\end{figure}
\subsection{The Instruction Used in Our Experiments}
In this section, we give three different kinds of instructions used in our experiments: the Long Instruction, the Short Instruction, and the Vanilla. 

As shown in figure~\ref{fig:longtoshort_prompt}, Long Instruction and Short Instruction are used in the Long-Short Sampling method to prompt the model to generate long and short reasoning trajectories, respectively. Vanilla refers to the instruction used by the model during training and inference, which does not employ any control over the length of the generated reasoning trajectory.

\end{document}